\documentclass[12pt, a4paper]{article}
\usepackage{amsmath,amsfonts,amsthm,mathrsfs}

\numberwithin{equation}{section}

\begin{document}

\title{Universality of Deep Convolutional Neural Networks}
\author{Ding-Xuan Zhou \\ Department of Mathematics, City University of Hong Kong \\
Kowloon, Hong Kong \\
Email: mazhou@cityu.edu.hk}
\date{}

\maketitle
\begin{abstract}
Deep learning has been widely applied and brought breakthroughs in speech recognition, computer vision, and many other domains.
Deep neural network architectures and computational issues have been well studied in machine learning.
But there lacks a theoretical foundation for understanding the approximation or generalization ability of deep learning methods generated by the network  architectures such as
deep convolutional neural networks. Here we show that a deep convolutional neural network (CNN) is universal, meaning that
it can be used to approximate any continuous function to an arbitrary accuracy
when the depth of the neural network is large enough. This answers an open question in learning theory.
Our quantitative estimate, given tightly in terms of the number of free parameters to be computed, verifies the efficiency of deep CNNs in dealing with large dimensional data.
Our study also demonstrates the role of convolutions in deep CNNs.
\end{abstract}


\section{Introduction and Main Results}\label{Sec.Introduction}

Deep learning provides various models and algorithms to process data as efficiently
as biological nervous systems or neuronal responses in the human brain \cite{Lecun1998, Hinton2006, Krizhevsky2012, Goodfellow2016, LecunNature}.
It is based on deep neural network architectures and those structures bring essential tools for obtaining data features and function representations in practical applications.
A main concern about deep learning which has attracted much scientific attention and some criticism is its lack of
theories supporting its practical efficiency caused by its network structures, though there have been some theoretical attempts
from approximation theory viewpoints \cite{Mallat, Mallat2016, Poggio}.
In particular, for deep CNNs having convolutional structures without fully connected layers, it is unknown which kinds of functions can be approximated.
This paper provides a rigorous mathematical theory to answer this question and to illustrate the role of convolutions.

\bigskip

\noindent {\bf Notation and Concepts}

\medskip

\noindent {\bf Convolutional Filters and Matrices.}

\medskip

The deep CNNs considered in this paper have two essential ingredients: a rectified linear unit (ReLU) defined as a univariate nonlinear function $\sigma$ given by
$$\sigma (u) = (u)_+ = \max\{u, 0\}, \qquad u\in  {\mathbb R}$$
and a sequence of convolutional filter masks ${\bf w} =\{w^{(j)}\}_j$
inducing sparse convolutional structures. Here a filter mask $w=(w_k)_{k=-\infty}^{\infty}$ means a sequence of filter coefficients.
We use a fixed integer filter length $s \geq 2$ to control the sparsity, and assume that
$w^{(j)}_k \not=0$ only for $0\leq k \leq s$.
The convolution of such a filter mask $w$ with another sequence $v =(v_0, \ldots,v_D)$ is a sequence $w{*}v$ given by
$\left(w{*}v\right)_i = \sum_{k=0}^{D} w_{i-k} v_k$. This leads to
a $(D+s) \times D$ Toeplitz type convolutional matrix $T$
which has constant diagonals:
$$
T = \left[\begin{array}{llllllll}
w_0 & 0 & 0 & 0 & \cdot & 0  \\
w_1 & w_0 & 0 & 0 & \cdot & 0  \\
\vdots & \ddots & \ddots & \ddots & \ddots & \vdots \\
w_{s} & w_{s-1} & \cdots & \hspace*{-3mm} w_0 &  \hspace*{-3mm}  0 \cdots & 0  \\
0 & w_s & \cdots &  \hspace*{-3mm} w_1 &  \hspace*{-3mm} w_0 \cdots \ & 0 \\
\vdots & \ddots  & \ddots \ddots & \ddots  & \ddots \ddots & \vdots   \\
\cdots  & \cdots & 0 & \hspace*{-4mm} w_{s} & \cdots & w_0   \\
\cdots  & \cdots & \cdots &  \hspace*{-4mm} 0 &  \hspace*{-6mm} w_{s} \cdots & w_1   \\
\vdots & \ddots & \ddots & \ddots & \ddots & \vdots   \\
0 & \cdots  & \cdots \cdots &   \cdots  0  & \hspace*{6mm}  w_{s}  &  w_{s-1} \\
0 & \cdots & \cdots & \cdots &  \hspace*{4mm} \cdots 0 & w_{s}
\end{array}\right].
$$
Sparse matrices of this form induce deep CNNs which are essentially different from the classical neural networks involving full connection matrices.
Note that the number of rows of $T$ is $s$ greater than that of columns. This leads us to take a sequence of  linearly increasing widths $\{d_j = d + j s\}$ for the network,
which enables the deep CNN to represent functions of richer structures.

\medskip

\noindent {\bf Convolutional Neural Networks.}

\medskip

Starting with the action of $T$ on the input data vector $x \in {\mathbb R}^d$,
we can define a deep CNN of depth $J$ as a sequence of $J$ vectors
$h^{(j)}(x)$ of functions on ${\mathbb R}^d$ given iteratively by $h^{(0)}(x)=x$ and
$$
h^{(j)} (x) = \sigma \left(T^{(j)} h^{(j-1)}(x)  - b^{(j)}\right), \qquad j=1, 2, \ldots, J,
$$
where $T^{(j)} =\left(w^{(j)}_{i-k}\right)$ is a $d_j \times d_{j-1}$ convolutional matrix,
$\sigma$ acts on vectors componentwise,
and ${\bf b}$ is a sequence of bias vectors $b^{(j)}$.

Except the last iteration, we take $b^{(j)}$ of the form $[b_1 \ \ldots b_{s} \ b_{s+1} \ b_{s+1} \ \ldots \ b_{s+1} \ b_{d_j-s+1} \ \ldots b_{d_j}]^T$
with the $d_j - 2s$ repeated components in the middle. The sparsity of $T^{(j)}$ and the special form of $b^{(j)}$
tell us that the $j$-th iteration of the deep CNN involves $3 s +2$ free parameters. So in addition to the $2 d_J + s +1$ free parameters for $b^{(J)}, c \in {\mathbb R}^{d_J}, w^{(J)}$,
the total number of free parameters in the deep CNN is $(5 s +2) J + 2d -2 s -1$, much smaller than
that in a classical fully connected multi-layer neural network with full connection matrices $T^{(j)}$ involving $d_j d_{j-1}$ free parameters. It demonstrates the computational efficiency of deep CNNs.

\bigskip

\noindent {\bf Mathematical Theory of Deep CNNs}

\medskip

The hypothesis space of a learning algorithm is the set of all possible functions that can be represented or produced by the algorithm.
For the deep CNN of depth $J$ considered here, the hypothesis space is a set of functions defined by
$${\mathcal H}^{{\bf w}, {\bf b}}_J =\left\{\sum_{k=1}^{d_{J}} c_k
h^{(J)}_k (x): c \in {\mathbb R}^{d_J}\right\}.$$
This hypothesis space and its approximation ability depend completely on the sequence of convolutional filter masks ${\bf w} =\{w^{(j)}\}_{j=1}^J$ and the sequence of bias vectors ${\bf b}=\{b^{(j)}\}_{j=1}^J$.
Observe that each function in the hypothesis space ${\mathcal H}^{{\bf w}, {\bf b}}_J$
is a continuous piecewise linear function (linear spline) on any compact subset $\Omega$ of ${\mathbb R}^d$.
Our first main result verifies the universality of deep CNNs, asserting that any function $f\in C(\Omega)$,
the space of continuous functions on $\Omega$ with norm $\|f\|_{C(\Omega)} =\sup_{x\in \Omega} |f(x)|$,
can be approximated by ${\mathcal H}^{{\bf w}, {\bf b}}_J$ to an arbitrary accuracy when the depth $J$ is large enough.

\medskip

\noindent {\bf Theorem A.}
Let $2 \leq s\leq d$. For any compact subset $\Omega$ of ${\mathbb R}^d$ and any $f\in C(\Omega)$, there exist sequences ${\bf w}$ of filter masks, ${\bf b}$ of bias vectors and $f^{{\bf w}, {\bf b}}_J \in {\mathcal H}^{{\bf w}, {\bf b}}_J$ such that
$$
\lim_{J\to\infty} \|f-f^{{\bf w}, {\bf b}}_J \|_{C(\Omega)} = 0.
$$

Our second main result presents rates of approximation by deep CNNs for functions in the Sobolev space $H^{r} (\Omega)$ with an integer index $r>2 + d/2$.
Such a function $f$ is the restriction to $\Omega$ of a function $F$ from the Sobolev space $H^{r} ({\mathbb R}^d)$ on ${\mathbb R}^d$ meaning that $F$ and all its partial derivatives up to order $r$ are square integrable on ${\mathbb R}^d$.

\medskip

\noindent{\bf Theorem B.}
Let $2 \leq s\leq d$ and $\Omega \subseteq [-1, 1]^d$. If $J \geq 2d/(s-1)$ and $f=F|_{\Omega}$ with $F \in H^{r} ({\mathbb R}^d)$ and an integer index $r>2 + d/2$,
then there exist ${\bf w}$, ${\bf b}$ and $f^{{\bf w}, {\bf b}}_J \in {\mathcal H}^{{\bf w}, {\bf b}}_J$ such that
$$\|f-f^{{\bf w}, {\bf b}}_J \|_{C(\Omega)} \leq c \left\|F\right\| \sqrt{\log J} \left(1/J\right)^{\frac{1}{2} + \frac{1}{d}},$$
where $c$ is an absolute constant and $\left\|F\right\|$ denotes the Sobolev norm of $F \in H^{r} ({\mathbb R}^d)$.

\medskip

According to Theorem B, if we take $s=\lceil 1+ d^\tau/2\rceil$ and $J= \lceil 4 d^{1-\tau}\rceil L$ with $0 \leq \tau \leq 1$ and $L \in {\mathbb N}$,
where $\lceil u\rceil$ denotes the smallest integer not smaller than $u$, then we have
$$ \|f-f^{{\bf w}, {\bf b}}_J \|_{C(\Omega)} \leq c \left\|F\right\| \sqrt{\frac{(1-\tau) \log d + \log L + \log 5}{4 d^{1-\tau} L}},$$
while the widths of the deep CNN are bounded by $12 L d$ and the total number of free parameters by
$$(5 s +2) J + 2d -2 s -1 \leq (73 L +2) d. $$
We can even take $L=1$ and $\tau=1/2$ to get a bound for the relative error
$$ \frac{\|f-f^{{\bf w}, {\bf b}}_J \|_{C(\Omega)}}{\left\|F\right\|} \leq \frac{c}{2}  d^{-\frac{1}{4}} \sqrt{\log (5\sqrt{d})} $$
achieved by a deep CNN of depth $\lceil 4 \sqrt{d}\rceil$ and at most $75 d$ free parameters, which decreases as the dimension $d$ increases.
This interesting observation is new for deep CNNs, and does not exist in the literature of fully connected neural networks.
It explains the strong approximation ability of deep CNNs.

A key contribution in our theory of deep CNNs is that
an arbitrary pre-assigned sequence $W=(W_k)_{-\infty}^{\infty}$ supported in $\{0, \ldots, {\mathcal M}\}$
can be factorized into convolutions of a mask sequence $\{w^{(j)}\}_{j=1}^{J}$.
It is proved by the same argument as in \cite{ZhouAA} for the case with the special restriction $W_0 \not=0$.
Convolutions are closely related to translation-invariance in speeches and images \cite{Mallat, Daubechies, Strang}, and also in some learning algorithms \cite{SmaleZhou, FanHuWuZhou}.

\medskip

\noindent{\bf Theorem C.}
Let $s \geq 2$ and $W=(W_k)_{-\infty}^{\infty}$ be a sequence supported in $\{0, \ldots, {\mathcal M}\}$ with ${\mathcal M} \geq 0$.
Then there exists a finite sequence of filter masks $\{w^{(j)}\}_{j=1}^{J}$ supported in $\{0, \ldots, s\}$ with $J < \frac{{\mathcal M}}{s-1}+1$
such that the convolutional factorization $W  = w^{(J)}{*}\ldots {*}w^{(2)}{*}w^{(1)}$ holds true.

\bigskip

\noindent{\bf Discussion}

\medskip

The classical shallow neural networks associated with an activation function $\sigma: {\mathbb R}  \to {\mathbb R}$ produce functions of the form
$$
f_N (x) =\sum_{k=1}^N c_k \sigma(\alpha_k \cdot x - b_k)
$$
with $\alpha_k \in {\mathbb R}^d, b_k, c_k \in {\mathbb R}$. A mathematical theory for approximation of functions by shallow neural networks was well developed three decades ago  \cite{Cybenko, Hornik, Barron, Mhaskar1993, Leshno1993, Pinkus1999}
and was extended to fully connected multi-layer neural networks shortly afterwards  \cite{Hornik, Mhaskar1993, Chui1996} .

The first type of results obtained in the late 1980s  are about universality,
asserting that any continuous function $f$ on any compact subset $\Omega$ of ${\mathbb R}^d$ can be approximated by some $f_N$ to an arbitrary accuracy when the number of hidden neurons $N$ is large enough.
Such results were given in  \cite{Cybenko, Hornik, Barron} when $\sigma$ is a sigmoidal function, meaning that $\sigma$ is a continuous strictly increasing function
satisfying $\lim_{u\to-\infty}\sigma(u)=0$ and $\lim_{u\to \infty}\sigma(u)=1$. A more general result with a locally bounded and piecewise continuous activation function $\sigma$
asserts \cite{Leshno1993, Pinkus1999} that universality holds if and only if $\sigma$ is not a polynomial.

The second type of results obtained in the early 1990s are about rates of approximation. When $\sigma$ is a $C^\infty$ sigmoidal function
and $f=F|_{[-1, 1]^d}$ for some $F\in L^2({\mathbb R}^d)$ with the Fourier transform $\hat F$  satisfying $|w| \hat F(w) \in L^1 ({\mathbb R}^d)$,
rates of type $\|f_N - f\|_{L^2_{\mu} ([-1, 1]^d)} =O(1/\sqrt{N})$  were given in \cite{Barron} where $\mu$ is an arbitrary probability measure $\mu$.
Analysis was conducted in \cite{Mhaskar1993} for shallow neural networks with more general continuous activation functions $\sigma$
satisfying a special condition with some $b\in{\mathbb R}$ that $\sigma^{(k)} (b)\not= 0$ for any nonnegative integer $k$ and a further assumption with some integer $\ell \not=1$ that
$\lim_{u\to-\infty} \sigma(u)/|u|^\ell=0$ and $\lim_{u\to \infty} \sigma(u)/u^\ell=1$.
The rates there are of type $\left\|f_N - f\right\|_{C([-1, 1]^d)} =O(N^{-r/d})$ for
$f \in C^r ([-1, 1]^d)$. Note that the ReLU activation function considered in this paper
does not satisfy the condition with $\sigma^{(k)} (b)\not= 0$ or the special assumption with $\ell\not= 1$.
To achieve the approximation accuracy $\left\|f_N - f\right\|_{C([-1, 1]^d)}  \leq \epsilon$,  when $r= \lceil \frac{d+1}{2} +2\rceil$ with $d/r \approx 2$,
the number of hidden neurons $N \geq \left(c_{f, d, \ell}/\epsilon\right)^{d/r}$ and the total number of free parameters is at least $\left(c_{f, d, \ell}/\epsilon\right)^{d/r} d$,
where the constant $ c_{f, d, \ell}$ depends on the dimension $d$ and might be very large.
To compare with our result, we take the filter length $s=\lceil 1+ d/2\rceil$ and depth $J= 4 L$ with $L \in {\mathbb N}$. We know from Theorem B that
the same approximation accuracy $ \|f-f^{{\bf w}, {\bf b}}_J \|_{C(\Omega)} \leq \epsilon$ with $0< \epsilon \leq c \left\|F\right\|$
can be achieved by the deep CNN of depth $J= 4 \lceil \frac{1}{\epsilon^2} \log  \frac{1}{\epsilon^2}\rceil$ having at most
$ \lceil \frac{75}{\epsilon^2} \log  \frac{1}{\epsilon^2}\rceil d$ free parameters, which does not depend on the dimension $d$.
Though a logarithmic term is involved, this dimension independence gives evidence for the power of deep CNNs.

A multi-layer neural network is a sequence of function vectors
$h^{(j)}(x)$ satisfying an iterative relation
$$
h^{(j)} (x) = \sigma \left(T^{(j)} h^{(j-1)}(x)  - b^{(j)}\right), \qquad j=1, 2, \ldots, J.
$$
Here $T^{(j)}$ is a full connection matrix without special structures.
So a deep CNN is a special multi-layer neural network with sparse convolutional matrices. This sparsity gives difficulty in developing a mathematical theory for deep CNNs,
since the techniques in the literature of fully connected shallow or multi-layer neural networks do not apply.
Our novelty to overcome the difficulty is
to factorize an arbitrary finitely supported sequence into convolutions of filter masks $\{w^{(j)}\}_{j=1}^J$ supported in $\{0, 1, \ldots, s\}$.
Our method can be applied to distributed learning algorithms \cite{LGZ, GLZ}.

Recently there have been quite a few papers \cite{Telgarsky2016, Eldan2016, Yarosky, Coifman, Grohs, Petersen}
on approximation and representation of functions by deep neural networks and benefit of depth, but all these results are for fully connected networks without pre-specified structures, not for deep CNNs.
In particular, it was shown in \cite{Grohs, Petersen} that the rate of approximaton of some function classes by multi-layer fully connected neural networks may be achieved by networks
with sparse connection matrices $T^{(j)}$, but the locations of the sparse connections are unknown. This sparsity of unknown pattern is totally different from that of deep CNNs, the latter enables
computing methods like stochastic gradient descent to learn values of the free parameters efficiently.

Deep CNNs are often combined with pooling, a small number of fully connected layers, and some other techniques for improving the practical performance of deep learning.
Our purpose to analyze purely convolutional networks is to demonstrate that convolution makes full use of shift-invariance properties of speeches and images for extracting data features efficiently.
Also, for processing an image, convolutions based on the 2-D lattice ${\mathbb Z}^2$ are implemented by taking inner products of $(s+1) \times (s+1)$ filter matrices with shifted patches of the image.
Though we do not consider such deep learning algorithms in this paper, some of our ideas can be used to establish mathematical theories for more general deep neural networks involving convolutions.

\bigskip

\noindent{\bf Methods}

\medskip

For approximation in $C(\Omega)$ we can only consider those Sobolev spaces which can be embedded into the space of continuous functions,
that is, those spaces with the regularity index $r > \frac{d}{2}$. To establish rates of approximation we require $r>\frac{d}{2} +2$ in Theorem B. In this case, the set $H^r (\Omega)$ is dense in $C(\Omega)$, so
Theorem A follows from Theorem B by scaling.

\bigskip

\noindent{\bf Proof of Theorem B.}
Let $J \geq \frac{2d}{s-1}$ and $m$ be the integer part of $\frac{(s-1)J}{d} -1\geq 1$. In our assumption, $f =F|_{\Omega}$ for some function $F \in H^{r} ({\mathbb R}^d)$ with the Fourier transform $\widehat{F}(\omega)$ giving the norm
$\left\|F\right\| =\left\|\left(1+ |\omega|^2\right)^{r/2} \widehat{F}(\omega)\right\|_{L^2}$.
By the Schwarz inequality and the condition $r>\frac{d}{2} +2$,
$v_{F, 2}:= \int_{{\mathbb R}^d} \|\omega\|_1^2 \left|\widehat{F}(\omega)\right| d\omega \leq c_{d, r} \left\|F\right\|$
where $c_{d, r}$ is the finite constant $\left\|\|\omega\|_1^2 \left(1+ |\omega|^2\right)^{-r/2}\right\|_{L^2}$.
Then we apply a recent result from \cite{Klusowski2018} on ridge approximation to $F|_{[-1, 1]^d}$ and know that there exists a linear combination of ramp ridge functions of the form
$$
F_m (x) = \beta_0 + \alpha_0 \cdot x + \frac{v}{m} \sum_{k=1}^m \beta_k \left(\alpha_k \cdot x - t_k\right)_+
$$
with $\beta_k \in [-1, 1], \|\alpha_k\|_1 =1, t_k \in [0, 1], \beta_0 = F(0), \alpha_0 = \nabla F(0)$ and $|v| \leq 2 v_{F, 2}$ such that
$$
\left\|F - F_m\right\|_{C([-1, 1]^d)} \leq c_0 v_{F, 2} \max\left\{\sqrt{\log m}, \sqrt{d}\right\} m^{-\frac{1}{2} - \frac{1}{d}}
$$
for some universal constant $c_0>0$.

Now we turn to the key step of constructing the filter mask sequence ${\bf w}$.
Define a sequence $W$ supported in $\{0, \ldots, (m+1) d-1\}$ by stacking the vectors $\alpha_0, \alpha_1, \ldots, \alpha_m$ (with components reversed) by
$$\left[W_{(m+1) d -1} \ \ldots \ W_{1} \ W_{0}\right] = \left[\alpha_m^T \ \ldots \ \alpha_1^T \ \alpha_0^T\right].$$
We apply Theorem C to the sequence $W$ with support in $\{0, 1, \ldots, (m+1) d\}$
and find a sequence of filter masks ${\bf w}=\{w^{(j)}\}_{j=1}^{\hat{J}}$ supported in $\{0, 1, \ldots, s\}$
with $\hat{J} < \frac{(m+1) d}{s-1}+1$ such that $W  = w^{(\hat{J})}{*}w^{(\hat{J}-1)}{*}\ldots {*}w^{(2)}{*}w^{(1)}.$
The choice of $m$ implies $\frac{(m+1) d}{s-1} \leq J$. So $\hat{J} \leq J$ and by taking $w^{(\hat{J}+1)} = \ldots = w^{(J)}$ to be the delta sequence, we have
$W  = w^{(J)}{*}w^{(J-1)}{*}\ldots {*}w^{(2)}{*}w^{(1)}.$
This tells us \cite{ZhouAA} that
$$T^{(J)} \ldots T^{(1)}= T^{W}$$
where
$T^{W}$ is the $d_J \times d$ matrix given by $[W_{\ell -k}]_{\ell=1, \ldots, d_J, k=1, \ldots, d}$. Observe
from the definition of the sequence $W$ that for $k =0, 1, \ldots, m$, the $(k+1)d$-th row of $T^{W}$ is exactly the transpose of $\alpha_k$.
Also, since $Js \geq (m+1) d$, we have $W_{Js}=0$ and the last row of $T^{W}$ is a zero row.

Then we construct ${\bf b}$. Denote $\|w\|_1=\sum_{k=-\infty}^{\infty} |w_k|$,
$B^{(0)} =\max_{x\in \Omega} \max_{k=1, \ldots, d} |x_k|$ and
$B^{(j)} = \|w^{(j)}\|_1 \ldots \|w^{(1)}\|_1 B^{(0)}$ for $j\geq 1$. Then we have
$$\left\|\left(T^{(j)} \ldots T^{(1)}  x\right)_k\right\|_{C(\Omega)} \leq B^{(j)}, \qquad \forall k=1, \ldots, d_j.$$
Take $b^{(1)} =- B^{(1)} {\bf 1}_{d_1} :=- B^{(1)} (1, \ldots, 1)^T$, and
$$b^{(j)} =
B^{(j-1)} T^{(j)} {\bf 1}_{d_{j-1}} - B^{(j)} {\bf 1}_{d_j}, \qquad j=1, \ldots, J-1. $$
Then for $j=1, \ldots, J-1$, we have
$$
h^{(j)} (x) = T^{(j)}\ldots T^{(1)} x + B^{(j)} {\bf 1}_{d_j}$$
and $b^{(j)}_{\ell} = B^{(j-1)} \sum_{k=0}^{s} w^{(j)}_{k} - B^{(j)}= b^{(j)}_{s+1}$ for $\ell=s+1, \ldots, d_j -s.$ Hence the bias vectors are of the required form.

Finally, we take the bias vector $b^{(J)}$ by setting $b^{(J)}_{\ell}$ to be
$$ \left\{\begin{array}{ll}
B^{(J-1)} (T^{(J)} {\bf 1}_{d_{J-1}})_{\ell} - B^{(J)}, & \hbox{if} \ \ell =d, d + Js, \\
B^{(J-1)} (T^{(J)} {\bf 1}_{d_{J-1}})_{\ell} + t_k, & \hbox{if} \ \ell =(k+1)d, \ 1 \leq k \leq m, \\
 B^{(J-1)}(T^{(J)} {\bf 1}_{d_{J-1}})_{\ell} + B^{(J)}, & \hbox{otherwise.} \end{array}\right. $$
Substituting this bias vector and the expression for $h^{(J-1)} (x)$ into the iterative relation of the deep CNN,
we see from the identity $T^{(J)} \ldots T^{(1)}= T^{W}$ and the definition of the sequence $W$ that the $\ell$-th component
$h^{(J)}_{\ell} (x)$ of $h^{(J)} (x)$ equals
$$ \left\{\begin{array}{ll}
\alpha_0 \cdot x + B^{(J)}, & \hbox{if} \ \ell =d,  \\
 B^{(J)}, & \hbox{if} \ \ell =d +Js,  \\
\left(\alpha_k \cdot x - t_k\right)_+, & \hbox{if} \ \ell =(k+1)d, \ 1 \leq k \leq m, \\
0, & \hbox{otherwise.} \end{array}\right. $$
Thus, we can take
$f^{{\bf w}, {\bf b}}_J = F_m|_{\Omega} \in \hbox{span}\{h^{(J)}_{k} (x)\}_{k=1}^{d_J} ={\mathcal H}^{{\bf w}, {\bf b}}_J$ and know that the error
$\|f-f^{{\bf w}, {\bf b}}_J \|_{C(\Omega)} \leq \left\|F - F_m\right\|_{C([-1, 1]^d)}$ can be bounded as
$$\|f-f^{{\bf w}, {\bf b}}_J \|_{C(\Omega)} \leq c_0 v_{F, 2} \max\left\{\sqrt{\log m}, \sqrt{d}\right\} m^{-\frac{1}{2} - \frac{1}{d}}.
$$
But $\frac{1}{2} (s-1)J \leq m d < (s-1)J$ and $2 r -d -4 \geq 1$. By a polar coordinate transformation,
$c_{d, r} d^{1 + \frac{1}{d}}
\leq \sqrt{\frac{d^6 \pi^{d/2}}{\Gamma (\frac{d}{2} +1)}} \left(1 + \frac{1}{\sqrt{2 r -d -4}}\right)$ which can be bounded by an absolute constant
 $c' :=\max_{\ell\in {\mathbb N}} 2\sqrt{\ell^6 \pi^{\ell/2}/\Gamma (\frac{\ell}{2} +1)}$. Therefore,
$$\|f-f^{{\bf w}, {\bf b}}_J \|_{C(\Omega)} \leq 2 c_0 c' \left\|F\right\| \sqrt{\log J} J^{-\frac{1}{2} - \frac{1}{d}}.
$$ This proves Theorem B by taking $c = 2 c_0 c'$.

\bigskip

Convolutional factorizations have been considered in our recent work \cite{ZhouAA} for sequences $W$ supported in $\{0, 1, \ldots, S\}$ with $S \geq d$ under the special restrictions $W_0>0$ and $W_{S}\not= 0$.
Theorem C gives a more general result by improving the bound for $J$ in \cite{ZhouAA} and removing the special restrictions on $W_0$ and $W_S$.

\bigskip

\noindent{\bf Proof of Theorem C.}
We apply a useful concept from the literature of wavelets \cite{Daubechies}, the symbol $\widetilde{w}$ of a sequence $w$ finitely supported in the set of nonnegative integers, defined as a polynomial on ${\mathbb C}$ by $\widetilde{w} (z) = \sum_{k=0}^\infty w_k z^k$.
The symbol of the convoluted sequence $a{*}b$ is given by $\widetilde{a{*}b} (z) = \widetilde{a} (z)\widetilde{b} (z)$.
Notice that the symbol $\widetilde{W}$ of  the sequence $W$ supported in $\{0, \ldots, {\mathcal M}\}$  is a polynomial of degree $M$ with real coefficients for some $0 \leq M \leq {\mathcal M}$.
So we know that complex roots $z_k =x_k + i y_k$ of $\widetilde{W}$ with $x_k \not=0$ appear in pairs
and by $(z-z_k)(z- \overline{z_k}) =z^2 - 2 x_k  z + \left(x_k^2 + y_k^2\right)$, the polynomial $\widetilde{W} (z)$ can be completely factorized as
$$
\widetilde{W} (z) =W_{M} \Pi_{k=1}^K \left\{z^2 - 2 x_k  z + \left(x_k^2 + y_k^2\right)\right\} \Pi_{k=2K+1}^{M} (z-x_k),
$$
where $2K$ is the number of complex roots with multiplicity, and $M - 2 K$ is the number of real roots with multiplicity.
By taking groups of up to $s/2$ quadratic factors (or $(s-1)/2$ quadratic factors with a linear factor) and $s$ linear factors in the above factorization,
we get $\widetilde{W} (z) = \widetilde{w^{(J)}}(z) \ldots \widetilde{w^{(2)}}(z) \widetilde{w^{(1)}}(z)$,  a factorization of $\widetilde{W}$ into polynomials of degree up to $s$, which yields a desired convolutional factorization $W  = w^{(J)}{*}w^{(J-1)}{*}\ldots {*}w^{(2)}{*}w^{(1)}$ and proves Theorem C.

\section*{Acknowledgments}
The author would like to thank Gilbert Strang and Steve Smale for their detailed suggestions and encouragement.
The work described in this paper is supported partially by the Research Grants Council of Hong Kong [Project No CityU 11306617] and by National Nature Science Foundation of China [Grant No 11461161006].

\bibliographystyle{abbrvnat}

\end{document}